\documentclass[letterpaper 10pt, conference]{IEEEtran}

\IEEEoverridecommandlockouts

\usepackage[T1]{fontenc}
\usepackage{lmodern}
\usepackage{ae, aecompl}

\usepackage{subcaption}
\usepackage{hyperref}       
\usepackage{booktabs}       
\usepackage{placeins}
\usepackage[compress]{cite}
\usepackage{graphicx}
\usepackage[acronym,shortcuts]{glossaries}
\usepackage{comment}
\usepackage{enumerate}
\usepackage[capitalise]{cleveref}
\usepackage[table,xcdraw,dvipsnames]{xcolor}
\usepackage{siunitx}
\usepackage{bm} 
\usepackage{amssymb} 
\usepackage{amsmath}
\usepackage[english]{babel}
\usepackage{multirow}
\usepackage{pifont}
\usepackage{algorithm}
\usepackage{algpseudocode}
\usepackage{cuted}
\usepackage{soul}

\definecolor{lightgray}{gray}{0.9}

\def\BibTeX{{\rm B\kern-.05em{\sc i\kern-.025em b}\kern-.08em
    T\kern-.1667em\lower.7ex\hbox{E}\kern-.125emX}}
\newcommand{\rev}[1]{\textcolor{black}{#1}}
\usepackage[normalem]{ulem} 
\newcommand{\revdel}[1]{\textcolor{blue}{}}

\usepackage{eso-pic}

\newcommand\AtPageUpperMyleft[1]{\AtPageUpperLeft{%
\put(\LenToUnit{1cm},\LenToUnit{-2cm}){#1}%
}}%

\AddToShipoutPictureBG*{%
  \AtPageUpperMyleft{\parbox[b][2cm][c]{\paperwidth}{%
    \centering
    \fontsize{12}{14}\selectfont
    \color{gray!50}
    This paper has been accepted for publication at\\
    IEEE Robotics and Automation Letters (RA-L) \copyright{}IEEE.
  }}%
}

\AddToShipoutPictureBG*{
  \AtPageLowerLeft{%
    \raisebox{20pt}{\makebox[\paperwidth]{\begin{minipage}{21cm}\centering
    \fontsize{10}{10}\selectfont
      \textcolor{gray!50}{ \copyright 2024 IEEE.  Personal use of this material is permitted.  Permission from IEEE must be obtained for all other uses, in any current or future media, including reprinting/republishing this material for advertising or promotional purposes, creating new collective works, for resale or redistribution to servers or lists, or reuse of any copyrighted component of this work in other works.
      }
    \end{minipage}}}%
  }
}

\title{\LARGE \bf \vspace{6mm}
Predictive Spliner: Data-Driven Overtaking in Autonomous Racing Using Opponent Trajectory Prediction
}

\author{Nicolas Baumann\IEEEauthorrefmark{1}, Edoardo Ghignone\IEEEauthorrefmark{1}, Cheng Hu\IEEEauthorrefmark{2}, Benedict Hildisch\IEEEauthorrefmark{1},\\ Tino Hämmerle\IEEEauthorrefmark{1}, Alessandro Bettoni\IEEEauthorrefmark{1}, Andrea Carron \IEEEauthorrefmark{3}, Lei Xie \IEEEauthorrefmark{2}, and Michele Magno\IEEEauthorrefmark{1} \\
\thanks{\IEEEauthorrefmark{1}Nicolas Baumann, Edoardo Ghignone, Benedict Hildisch, Tino Hämmerle, Alessandro Bettoni, and Michele Magno are associated with the Center for Project-Based Learning, D-ITET, ETH Zürich.}%
\thanks{\IEEEauthorrefmark{2}Cheng Hu and Lei Xie are associated with the Department of Control Science and Engineering, Zhejiang University.
}
\thanks{\IEEEauthorrefmark{3}Andrea Carron is associated with the Institute for Dynamic Systems and Control (IDSC), ETH Zürich.}
\thanks{\emph{(Nicolas Baumann, Edoardo Ghignone and Cheng Hu contributed equally to this work.) (Corresponding author: Nicolas Baumann.)}
}
}

\begin{document}
\newacronym{lidar}{LiDAR}{Light Detection and Ranging}
\newacronym{radar}{RADAR}{Radio Detection and Ranging}
\newacronym{mpc}{MPC}{Model Predictive Control}
\newacronym{mpcc}{MPCC}{Model Predictive Contouring Control}
\newacronym{iot}{IoT}{Internet of Things}
\newacronym{bev}{BEV}{Bird's-Eye View}
\newacronym{sota}{SotA}{State-of-the-Art}
\newacronym{gpu}{GPU}{Graphics Processing Unit}
\newacronym{fps}{FPS}{Frames Per Second}
\newacronym{kf}{KF}{Kalman Filter}
\newacronym{ads}{ADS}{Autonomous Driving Systems}
\newacronym{iac}{IAC}{Indy Autonomous Challenge}
\newacronym{fsd}{FSD}{Formula Student Driverless}
\newacronym{rrt*}{RRT*}{Rapidly exploring Random Tree Star}
\newacronym{mgbt}{MGBT}{Multilayer Graph-Based Trajectory}
\newacronym{ftg}{FTG}{Follow The Gap}
\newacronym{gbo}{GBO}{Graph-Based Overtake}
\newacronym{gp}{GP}{Gaussian Process}
\newacronym{rbf}{RBF}{Radial Basis Function}
\newacronym{map}{MAP}{Model- and Acceleration-based Pursuit}
\newacronym{sqp}{SQP}{Sequential Quadratic Programming}
\newacronym{roc}{RoC}{Region of Collision}
\newacronym{cpu}{CPU}{Central Processing Unit}
\newacronym{cots}{CotS}{Commercial off-the-Shelf}
\newacronym{obc}{OBC}{On-Board Computer}

\maketitle

\begin{strip}
\vspace{-3.55cm}
\centering
\includegraphics[width=\textwidth]{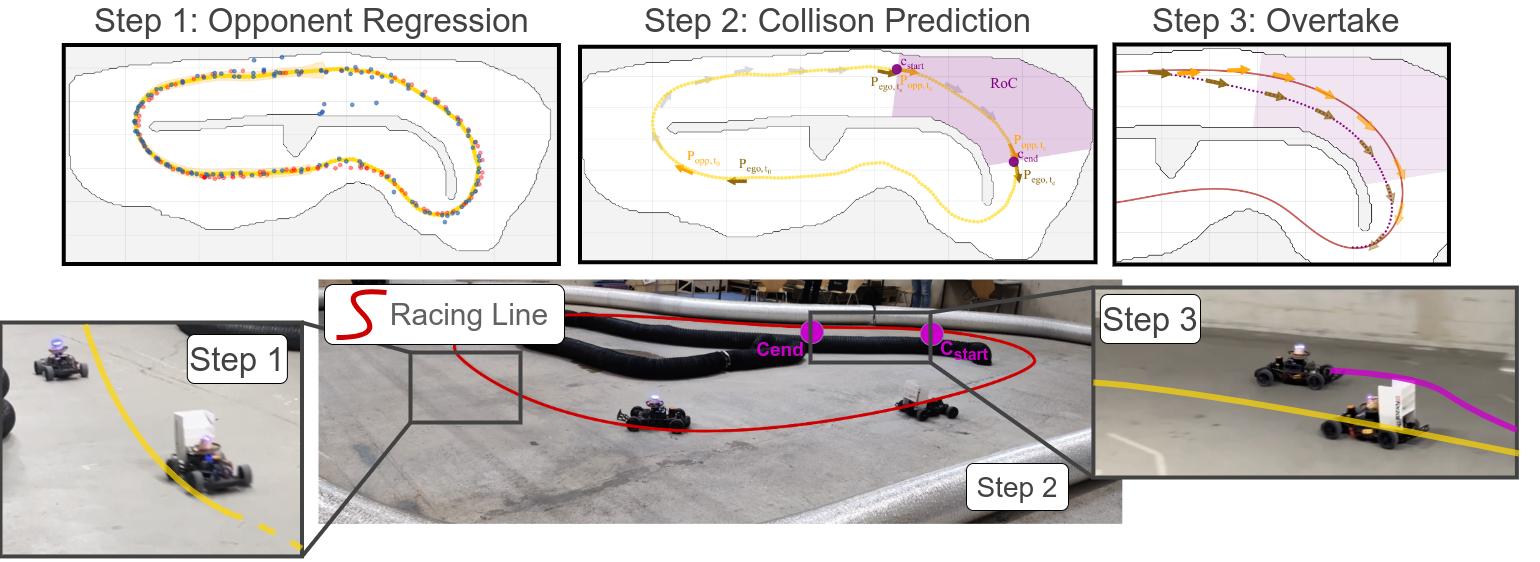}
\captionof{figure}{Qualitative visualization of the proposed \emph{Predictive Spliner} overtaking planner for autonomous racing. The method \rev{leverages} future opponent knowledge (\textcolor{Dandelion}{yellow}), initially regressing the opponent's trajectory via a \gls{gp} (Step 1). The \gls{roc}, defined by $c_{start}$ and $c_{end}$, is predicted by forward propagating the opponent's current and future poses through the \gls{gp} (Step 2). Finally, this knowledge is used to compute an overtaking trajectory in the future section of the track (\textcolor{purple}{purple}), avoiding premature \rev{overtaking} maneuvers and improving performance (Step 3).}
\label{fig:graphical_abstract}
\vspace{-0.275cm}
\end{strip}

\thispagestyle{empty}
\pagestyle{empty}

\begin{abstract}
Head-to-head racing against opponents is a challenging and emerging topic in the domain of autonomous racing. We propose Predictive Spliner, a data-driven overtaking planner \rev{designed to enhance competitive performance by anticipating opponent behavior. Using \gls{gp} regression, the method learns and predicts the opponent's trajectory, enabling the ego vehicle to calculate safe and effective overtaking maneuvers. Experimentally validated on a 1:10 scale autonomous racing platform, Predictive Spliner outperforms commonly employed overtaking algorithms by overtaking opponents at up to 83.1\% of its own speed, being on average 8.4\% faster than the previous best-performing method. Additionally, it achieves an average success rate of 84.5\%, which is 47.6\% higher than the previous best-performing method.}
 The \rev{proposed algorithm} maintains computational efficiency\revdel{ with a \gls{cpu} load of 22.79\% and a computation time of 8.4ms, evaluated on a \gls{cots} Intel i7-1165G7}, making it suitable for real-time robotic applications. These results highlight the potential of Predictive Spliner to enhance the performance and safety of autonomous racing vehicles. \rev{The code for Predictive Spliner is available at: \url{https://github.com/ForzaETH/predictive-spliner}.}
\end{abstract}

\setlength{\tabcolsep}{3pt}

\glsresetall

\section{Introduction}

Unrestricted multi-agent head-to-head racing, where the opponent is not restricted to remain on pre-defined racing lines, represents a highly interesting area of research in autonomous racing, whereas time-trials focuses on achieving the fastest lap-time on an unobstructed track \cite{raji2023erautopilot}. Head-to-head racing involves high-speed multi-robot interactions and remains a developing field within the autonomous racing research domain. By exposing the agent to extreme conditions and requiring complex multi-agent robotic interactions, this research underscores the necessity for further exploration in this area \cite{betz_weneed_ar, ar_survey, forzaeth}.

From a planning perspective, the objectives in racing involve two key aspects: computing a globally optimal racing line \cite{gb_optimizer}, typically performed offline, and reacting to opponents in head-to-head races through local planning \cite{forzaeth}. The local planner focuses on collision avoidance and executing overtaking maneuvers to maintain both competitiveness and safety in a dynamically changing environment \cite{forzaeth, ar_survey}.

Local planning techniques must be integrated within a holistic robotic framework, where an autonomous mobile robot adheres to the \emph{See-Think-Act} principle \cite{amr}, working together with other autonomy modules such as state-estimation, perception, and controls. Within autonomous racing, they play a vital role in supplying the local planner with the information needed about the ego agent's velocity and position, as well as the environmental context of the track layout and the opponent's position \cite{forzaeth}. 

The development of autonomous race cars presents numerous challenges in terms of embedded processing with constrained computational capacities \cite{forzaeth}. However, low computational time and real-time processing are crucial for high-speed applications \cite{mpcc, forzaeth}.

For an overtaking strategy, the goal is to develop a method that is robust and allows for operational simplicity, without requiring frequent adjustments to heuristics such as the opponent’s behavior. Yet, this simplicity often contrasts with the sensitivity to parameters and computational intensity required by model-based techniques \cite{mpcc}, as well as the detailed heuristics needed for reactive \cite{ftg, forzaeth}, sampling \cite{frenetplanner}, and graph-based \cite{gbo} overtaking algorithms. Moreover, these planning approaches generally focus only on the spatial positioning of the opponent, overlooking crucial temporal dynamics such as predicting the opponent’s future actions \cite{gbo, forzaeth, frenetplanner}. 

To address these limitations, we present \emph{Predictive Spliner} (see \Cref{fig:graphical_abstract}), a method that learns the opponent's behavior during the race in a data-driven approach with a \gls{gp} regressing a smooth opponent estimation derived from noisy raw \gls{lidar} observations of the opponent \cite{forzaeth}. This knowledge is then utilized to predict overtaking opportunities, integrating both spatial and temporal elements of the race environment without the need for complex heuristic adjustments. This approach represents a computationally efficient planning process that adapts to changes in both the track layout and opponent behavior.

To summarize the contributions of this work:
\begin{enumerate}[I]
    \item \textbf{Predictive Spliner:} We introduce a novel data-driven overtaking algorithm for autonomous racing, which leverages the learned opponent behavior and incorporates both spatial and temporal information in a \rev{computationally} lightweight\revdel{ and performant} manner which outperforms conventional planners by 47.6\% in terms of \rev{the} ratio of overtakes to crashes ($\mathcal{R}_{ot/c}$) with a low computational time of \SI{8.4}{\milli \second}, evaluated on a \gls{cots} Intel i7-1165G7.
    \item \textbf{Comparison:} The proposed algorithm is tested on a 1:10 scaled autonomous racing car against the most commonly employed overtaking algorithms in autonomous racing. The platform consists of fully \gls{cots} hardware such as a Hokuyo UST-LX10 \gls{lidar} sensor and an Intel NUC 10 \gls{obc}, which facilitates accessibility and reproducibility. 
    \item \textbf{Open-Source:} The proposed \emph{Predictive Spliner}, alongside the benchmarked overtaking algorithms, are fully incorporated in a \gls{cots} open-source full-stack implementation \cite{forzaeth}, to enhance reproducibility. 
\end{enumerate}

\section{Related Work} \label{sec:rw}

Classically, when comparing trajectory-generating local planning methods of either general autonomous driving or autonomous racing, one can coarsely categorize the different methods into the \rev{three}\revdel{four} subsequently elaborated strategies.

\subsection{Reactive Planners}
Within scaled autonomous racing, the \gls{ftg} method \cite{ftg} stands out as a widely used overtaking strategy. This purely reactive framework navigates by steering towards the largest gap detected by \gls{lidar} measurements, functioning effectively in both time-trials and head-to-head scenarios (i.e. it does not rely on state estimation, global planning, etc.). However, this inherently bounds the performance and flexibility of this algorithm, making it ill-suited for competitive racing \cite{forzaeth}.

In contrast, the \emph{Spliner} method \cite{forzaeth} fully adheres to the \emph{See-Think-Act} principle by computing a single evasion trajectory within the \emph{Frenet} frame \cite{forzaeth}. It uses a simple 1D spline fitted to the racing line with predefined heuristics to avoid opponents effectively.

While these methods are computationally lightweight, their reliance on heuristics denies optimality and thus limits the performance of the overall system especially at high speeds. However, even with their simplicity, these methods tend to be easy to tune and work surprisingly well but have an upper bound on their performance. \textit{Predictive Spliner} addresses these limitations by incorporating future spatiotemporal information of the opponent, leading to more strategic and efficient overtaking maneuvers without the reliance of excessive heuristics on the assumptions of the opponent's behavior.

\subsection{Sampling Based Path-Planning}
Sampling-based planning methods extend beyond reactive methods by utilizing heuristics and a cost function to generate and evaluate multiple evasion trajectories. The \emph{Frenet Planner} \cite{frenetplanner}, originally developed for highway lane-switching, exemplifies this by producing multiple trajectories within the \emph{Frenet} frame. Each trajectory usually carries a cost based on deviations from the racing line and proximity to track boundaries.

Graph-based sampling, another type of planner, structures the planning space using a lattice that enables graph-based search such as \cite{rrtstar, gbo}. A prominent racing example is the \gls{gbo} algorithm presented in \cite{gbo}\rev{, which can even be extended to consider game theoretic properties of the opponent \cite{game_graphs}}. Planners of this type convert the overtaking challenge into a graph-search problem, assigning high costs to paths near opponents or track boundaries. However, the decomposition of the overtaking problem into a graph formulation necessitates heuristics to determine the physical feasibility of reaching specific nodes, as well as the design of the cost function --- further, the extensive computations involved in sampling demand parallel processing capabilities to handle the substantial computational requirements.

\textit{Predictive Spliner} mitigates these issues by using \gls{gp} regression to predict opponent trajectories, reducing computational overhead and eliminating the need for complex heuristics.

\subsection{Model Based Overtaking}
In contrast to reactive and sampling-based planners, model-based overtaking strategies, primarily through \gls{mpc} techniques, leverage the physical dynamics of the racing car to optimize control and planning concurrently over a receding horizon. The \gls{mpcc} \cite{mpcc}, a prominent example, enhances its optimization by maximizing progress along the track's centerline and adapting boundaries to include opponents, thus enabling dynamic overtaking without a pre-computed optimal line. Although \gls{mpcc} is effective, hierarchical \gls{mpc} strategies \cite{raji2022motion, vazquez2020optimization} are considered superior to achieve faster lap times, due to their globally optimized trajectory planning. However, the general concept of spatially incorporating the opponent into the boundary constraints can still be applied. \rev{Further, game-theoretic approaches have been combined with \gls{mpc} \cite{game_mpc, game_mpc2}, where the \gls{mpc} constraints can act as collision constraints.}

Despite their high performance, \gls{mpc} methods are highly sensitive to the accuracy of the vehicle's model dynamics, particularly the notoriously difficult-to-model tire dynamics, which can vary significantly due to changes in track conditions like surface friction \cite{story_of_modelmismatch}. Consequently, while model-based methods offer a performant approach to autonomous racing, their practical implementation is challenging due to computational demands and the need for precise model parameters \cite{forzaeth}.

\textit{Predictive Spliner} addresses these challenges by focusing solely on planning without relying on model dynamics for vehicle control, thereby reducing sensitivity to model inaccuracies and varying track conditions. Additionally, the concept of leveraging opponent trajectory predictions can be integrated into an \gls{mpc} controller, enhancing its predictiveness and versatility. This is in fact demonstrated later in \Cref{tab:phys_res}.  \rev{Finally, although this work does not directly apply game-theoretic \gls{sota} strategies like those proposed in \cite{game_graphs, game_mpc, game_mpc2}, we believe that future work including a game-theoretic approach could benefit from the opponent modeling introduced here.}

\subsection{Summary of Existing Overtaking Algorithms}
\rev{The review of previous methods shows that each algorithm has distinct advantages and limitations, with fundamentally different approaches that prevent mutual enhancement, as in \cref{tab:overtaking_algorithms}. Moreover, prior methods focus solely on the opponent's spatial position, neglecting spatiotemporal planning and motion forecasting. To address these gaps, we introduce \textit{Predictive Spliner}, which incorporates future spatiotemporal information, building on the simple \emph{Spliner} architecture \cite{forzaeth}. The proposed algorithm balances high performance with minimal reliance on opponent behavior heuristics, incorporates physical boundary constraints, and requires low computational overhead. While our approach can enhance various planners, we demonstrate its effectiveness by applying it to both \emph{Spliner} and an \gls{mpc} framework.}

\begin{table}[!htb] 
    \centering 
    \resizebox{\columnwidth}{!}{%
    \begin{tabular}{l|c|c|c|c}
    \toprule
    \textbf{Algorithm} & \textbf{Temporal} & \textbf{Opp. Heuristics} & \textbf{Param. Sensitivity} & \textbf{Compute [\%]$\downarrow$} \\
    \midrule
    \gls{ftg} \cite{ftg} & No & High & \textbf{Low} & 25.2 \\
    Frenet \cite{frenetplanner} & No & High & \textbf{Low} & 36.7 \\
    Spliner \cite{forzaeth} & No & High & \textbf{Low} & \textbf{16.7} \\
    \gls{gbo} \cite{gbo} & No & High & High & 98.9 \\
    \gls{mpc} \cite{mpcc} & No & Low & High & 106.9 \\
    Pred. MPC \textbf{(ours)} & \textbf{Yes} & \textbf{Low} & High & 106.8  \\
    Pred. Spliner \textbf{(ours)} & \textbf{Yes} & \textbf{Low} & \textbf{Low} & 22.8  \\
    \bottomrule
    \end{tabular}%
    }
    \caption{Summary of existing overtaking strategies, denoting key characteristics of each method; \textit{Temporal} considerations of the opponent (desired yes); Reliance on \textit{Opponent Heuristics} due to opponent behavior (desired low); Susceptibility to \textit{Parameter Sensitivity} (desired low); Nominal \textit{Computational} load with \texttt{psutil}.}
    \label{tab:overtaking_algorithms}
\end{table}

\section{Methodology}
\rev{This section outlines the implementation of the \emph{Predictive Spliner} overtaking method in three parts: \cref{subsec:opp_traj} covers learning opponent behavior and trajectory via \gls{gp} regression; \cref{subsec:coll_pred} explains using this knowledge to compute the future \gls{roc}; and \cref{subsec:splining} details how the collision prediction is utilized to compute an effective overtaking trajectory in the future. This method builds on the \gls{cots} open-source racing stack from \cite{forzaeth}.}

\subsection{Opponent Trajectory Regression} \label{subsec:opp_traj}

The opponent is trailed for \rev{a single} lap on the racetrack to learn their spatial and velocity behavior, under the assumption that they will maintain similar patterns in proceeding laps. This regression process involves collecting the opponent's positional and velocity measurements, which are then binned into an array parametrized along the racing line\rev{'s} \( s \) \rev{coordinate} in \textit{Frenet} \rev{space}. Specifically, every 10 cm (denoted as $\Delta$), the observed spatial position \( d \) and longitudinal velocity \( v_s \) are recorded. Two separate \gls{gp} models are employed: one for \( d \) as a function of \( s \) using a \textit{Matérn} kernel \cite{gpbook}, and another for \( v_s \) as a function of \( s \) using an \gls{rbf} kernel \cite{gpbook}. The usage of the specific kernels in their respective spatial or velocity domain has been empirically determined, both in simulation as well as on the robotic system, to yield the highest accuracy. The algorithmic procedure is further elaborated in \cref{algo:opp_regression}.

\begin{algorithm}[b] 
\caption{Opponent Trajectory Regression} \label{algo:opp_regression}
\begin{algorithmic}[1]
\State \textbf{Input:} Observations \( \{(s_i, d_i, v_{s,i})\}_{i=1}^{N} \)
\State \textbf{Output:} Predicted lateral positions \( d(s) \) and velocities \( v_s(s) \)

\State \textbf{Data Collection:}
\State Collect spatial positions \( d \) and velocities \( v_s \) of the opponent at discrete points along the track.

\State \textbf{Binning:}
\For{each observation \( (s_i, d_i, v_{s,i}) \)}
    \State Compute the bin idx \( k = \left\lfloor \frac{s_i}{\Delta} \right\rfloor \), where \( \Delta = 0.1 \) m.
    \State Add \( (d_i, v_{s,i}) \) to bin \( \mathcal{B}_k \)
\EndFor

\State \textbf{Gaussian Process Regression:}
\State Define GP models: \( d(s) \sim \mathcal{GP}_d(\mu_d(s), k_d(s, s')) \) and \( v_s(s) \sim \mathcal{GP}_{v_s}(\mu_{v_s}(s), k_{v_s}(s, s')) \)

\State \textbf{Model Fitting:}
\State Fit the GP models on the binned data \( \mathcal{B} \)

\State \textbf{Return:} Fitted lateral positions \( d(s) \) and velocities \( v_s(s) \) from their respective \(\mathcal{GP}\)'s
\end{algorithmic}
\end{algorithm}

\begin{figure}[htb]
    \includegraphics[width=\columnwidth]{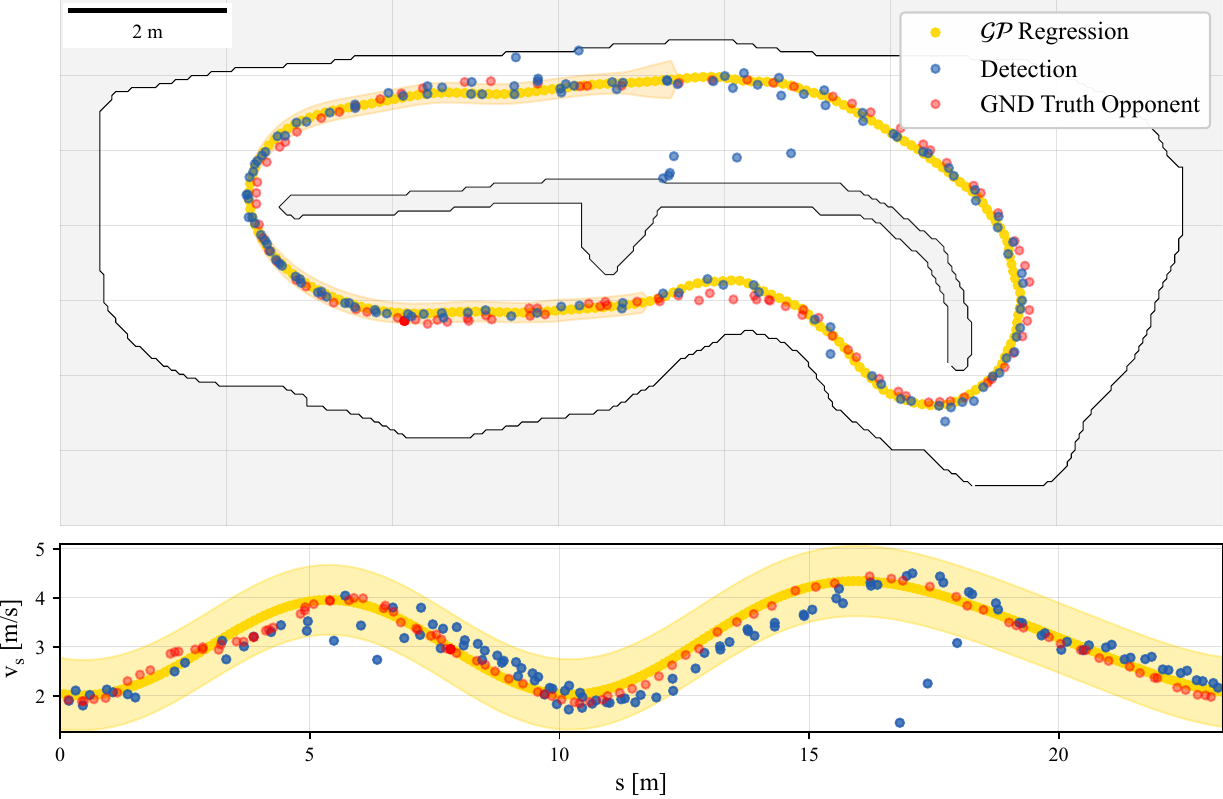}
    \caption{Visualisation of the regressed opponent trajectory on an example track, sampled from $\mathcal{GP}_{d,v_s}(s)$ in $d$ and $v_s$ in yellow with the standard deviations shaded. Opponent detections \texttt{obs} from the robotic perception module in blue. Ground truth opponent position and speed in red.}
    \label{fig: otexample}
\end{figure}

\Cref{fig: otexample} depicts an example of the opponent trajectory regression. Notice, how the \gls{gp}s allow for a smooth and continuous estimation of the opponent, as opposed to the sparse and noisy observed opponent measurements \texttt{obs} in blue. The covariance highlighted in yellow further allows for a simple state-machine trigger to abort the overtaking maneuver, in case the opponent was to behave unexpectedly\rev{, i.e. if the opponent's spatial or velocity behavior lies outside of the \gls{gp}'s covariance. Note, that the \gls{gp} regression can be used to continuously update the opponent's behavior on a per-lap basis.} The depicted information is then leveraged in the subsequent prediction stage, further outlined in \cref{subsec:coll_pred}.

\subsection{Collision Prediction} \label{subsec:coll_pred}
After the opponent's trajectory has been regressed, it is used to predict the point of collision. The prediction involves determining the starting ($c_{start}$) and ending \rev{(}$c_{end}$\rev{)} positions where a collision with the opponent is expected, parametrized over $s$ in \emph{Frenet} frame. For each time step, the opponent's speed at position \( s \) is sampled from the \( \mathcal{GP}_{v_s} \), and the corresponding \( s \) positions are computed. Specifically, the ego and opponent positions are forward propagated over a time horizon $\mathcal{T}_N$, to identify when a collision begins ($c_{start}$) and ends ($c_{end}$). This defines the future \gls{roc}\rev{, which is the track segment between $c_{start}$ and $c_{end}$. Specifically, $c_{start}$ marks the point where the front of the ego vehicle reaches the rear of the opponent, and $c_{end}$ is where the rear of the ego aligns with the front of the opponent, both obtained through forward propagation of their respective trajectories. This region represents the spatial overlap of both vehicles, assuming they could pass through each other.} \rev{To emphasize,} with the knowledge of the future \gls{roc}, the planner can anticipate potential collisions and determine an evasion trajectory in advance \rev{(i.e. leveraging spatial and temporal information of the opponent behavior)}, this enables the planner to delay overtaking maneuvers until they are safe and feasible. This approach is essential, as local planners that rely solely on spatial information may prematurely initiate overtaking maneuvers, resulting in deviations from the globally optimal racing line and thus a sacrifice in performance that will negatively impact overtaking capabilities.

Let \( s_{ego}(t) \) and \( s_{opp}(t) \) represent the positions of the ego and opponent vehicles at time \( t \). The forward propagation, as described in \cref{algo:coll_prediction}, utilizes the regressed \(\mathcal{GP}_{v_s}\) of the opponent's velocity to predict potential collision points along their future trajectories. It iteratively samples the opponent's velocity from \(\mathcal{GP}_{v_s}\) and updates both the ego and opponent positions over \(\mathcal{T}_N\). Collision points \(c_{start}\) and \(c_{end}\) are identified when the vehicles come within a collision threshold distance $\delta$, marking \(c_{start}\) and \(c_{end}\) when they then exceed it. 

\begin{figure}[htb]
    \centering
    \includegraphics[width=\columnwidth]{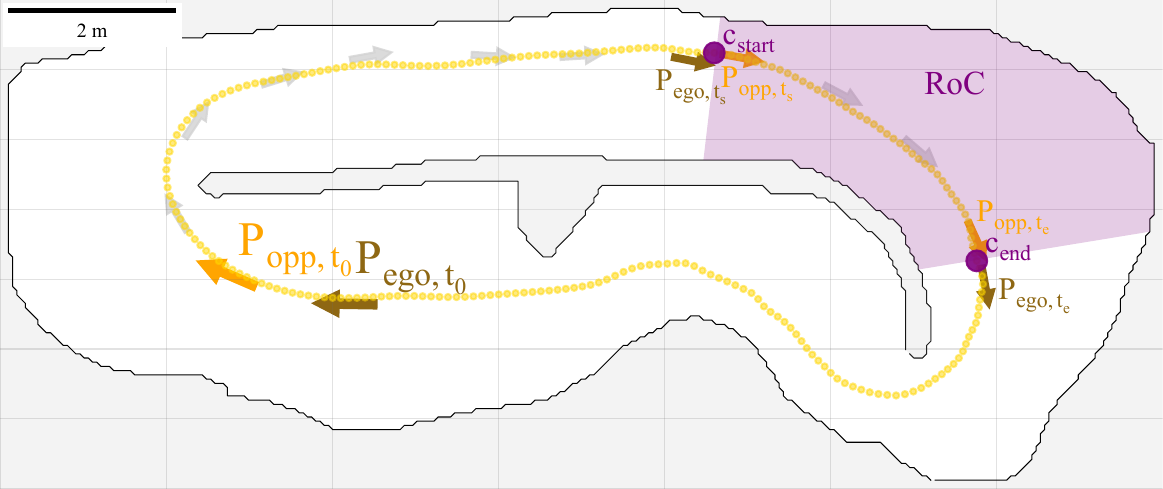}
    \caption{Example visualization of $c_{start,end}$ collision prediction. Current $t_0$ and future $t_{s,e}$ ego-agent poses ($P_{ego,t_i}$) in bronze, opponent poses ($P_{opp,t_i}$) in orange. Forward-propagated poses in grey, collision start ($c_{start}$) and end ($c_{end}$) at time $t_{s,e}$ in purple. The \gls{roc} is shaded in purple. In this case, the ego racing line and the opponent racing line both coincide with the optimal racing line.}
    \label{fig:collpred}
\end{figure}

\Cref{fig:collpred} illustrates this collision prediction algorithm, where the forward propagation of the current poses \(P_{ego,t_0}\) and \(P_{opp,t_0}\) over \(\mathcal{T}_N\) enables the computation of \(c_{start, end}\) at times \(t_{s, e}\) respectively, with the future predicted poses \(P_{ego,t_{s, e}}\) and \(P_{opp,t_{s, e}}\) spatially coinciding. Consequently, the \gls{roc} is defined as the racing track segment within the \(s\) coordinates of \(c_{start}\) and \(c_{end}\).

\begin{algorithm}[!b]
\caption{Collision Prediction}
\label{algo:coll_prediction}
\begin{algorithmic}[1]
\State \textbf{Input:} Regressed opponent trajectory \( \mathcal{GP}_{v_s}(s) \)
\State \textbf{Output:} Collision start \( c_{start} \) and end \( c_{end} \) positions
\State Initialize \( \Delta t, \delta, \text{ and } B \leftarrow \text{False} \)

\For{\( t \) in \( \mathcal{T}_N \)}
    \State Sample \( v_{opp}(t) \sim \mathcal{GP}_{v_s}(s_{opp}(t)) \)
    \State Compute new positions and speeds:
    \State \( s_{ego}(t + \Delta t) \leftarrow s_{ego}(t) + v_{ego}(t) \Delta t + \frac{1}{2} a_{ego}(t) (\Delta t)^2 \)
    \State \( s_{opp}(t + \Delta t) \leftarrow s_{opp}(t) + v_{opp}(t) \Delta t \)
    \State \( v_{ego}(t + \Delta t) \leftarrow v_{ego}(t) + a_{ego}(t) \Delta t \)
    \State Update \( c_{start} \) and \( c_{end} \) based on threshold \( \delta \)
    \If{\( \left| s_{opp}(t) - s_{ego}(t) \right| < \delta \) and not \( B \)}
        \State \( c_{start} \leftarrow s_{ego}(t) \), \( B \leftarrow \text{True} \)
    \ElsIf{\( \left| s_{opp}(t) - s_{ego}(t) \right| > \delta \) and \( B \)}
        \State \( c_{end} \leftarrow s_{ego}(t) \)
        \State \textbf{break}
    \EndIf
\EndFor
\end{algorithmic}
\end{algorithm}

\subsection{Optimization-based Spliner within the RoC} \label{subsec:splining}
As the \gls{roc} has now been computed, a simple \emph{Spliner} based spatial approach to overtaking as in \cite{forzaeth} can now be applied within the newly computed spatial boundaries of $c_{start/end}$ and the future knowledge of the opponent spatial behavior due to $\mathcal{GP}_d$. While the technique of leveraging the opponent's knowledge can enhance any overtaking strategy that relies solely on spatial information, as in \cref{tab:phys_res}, within this work we primarily demonstrate the effectiveness of leveraging opponent knowledge while utilizing a \gls{sqp} modified version of \emph{Spliner}.
The planner's objective is to adjust the d-coordinate of each ego-waypoint with the initial guess from the \emph{Spliner} while ensuring a safe distance away from the opponent's trajectory within \gls{roc}. The cost function of the algorithm can be formulated as follows:
\begin{align}
J(d^{ego}) = \sum^{N}_{i=0} Q_d \cdot d_i^{ego}+\sum^{N}_{i=1} Q_{d_s} \cdot \frac{\partial^2 d^{ego}_i}{\partial (s^{ego}_i)^2}+Q_{d_{\Delta}} \cdot \Delta_{d}^2
\end{align}
where $d^{ego}$\rev{/$s^{ego}$} represents the lateral\rev{/longitudinal} distance \rev{of the ego agent} from the global raceline. $N=\frac{\rev{\mathcal{T}_N}}{\Delta_t}$ is the number of points from $c_{start}$ to $c_{end}$. The step size $\Delta_{d}$ is defined as $d^{ego}_1-d^{ego}_0$. $Q_d$, $Q_{d_s}$, and $Q_{d_{\Delta}}$ are \rev{tuneable} weight parameters. The first term weighted with $Q_d$, aims to follow the avoidance trajectory as close as possible to the reference racing line. The second term weighted with $Q_{d_s}$, aims to ensure the trajectory's curvature smoothness by penalizing large second derivatives of the lateral displacement, which corresponds to sudden changes in lateral acceleration. The final term weighted with $Q_{d_{\Delta}}$, aims to ensure the trajectory is spatially smooth and does not exhibit sudden changes at the initial step. Lastly, it has been observed that the introduction of a penalty term for the first derivative of $d^{ego}_i$ did not yield substantial improvements, hence it was omitted for simplicity. From \cref{algo:coll_prediction}, $s_{opp}$ is used as the input of $\mathcal{GP}_d$ to predict the opponent's lateral distance $d^{opp}$ in the \gls{roc}. Then a safety constraint based on the future spatial knowledge of the opponent's car to avoid the collision is constructed as follows:
\begin{align}
\vert d^{ego}_i-d^{opp}_i\vert \geq \delta_{min} \label{eq:d_constr}
\end{align}
where $\delta_{min}$ encompasses the width of both cars with an additional safety margin. To generate a feasible and smooth trajectory, three constraints are also added.
The first is the turning radius constraint \eqref{eq: radius_constr}, which ensures the sum of the curvature from the \emph{Frenet} frame and the cartesian frame at each waypoint remains below a predefined threshold $\kappa_{max}$. 
\begin{align} 
\vert \kappa_{i,frenet}+\kappa_{i,global} \vert \leq \kappa_{max} \label{eq: radius_constr}
\end{align}
where $\kappa_{i,frenet}=\frac{\dot{s}_{ego} \ddot{d}_{ego}-\dot{d}_{ego} \ddot{s}_{ego}}{(\dot{s}_{ego}^2+\dot{d}_{ego}^2)^{2/3}}$, and $\kappa_{i,global}$ represents the curvature at the closest global waypoint to $s^{ego}_i$. This is necessary to ensure that the resulting trajectory adheres to the global reference curvature.
Furthermore, the track boundary constraint and terminal constraint are expressed as follows:
\begin{align}
&\delta_{r} \leq  d^{ego}_i \leq \delta_{l} \label{eq: bound_constr} \\ 
&\begin{bmatrix}d^{ego}_{N-1} &d^{ego}_{N}  \end{bmatrix}^{T}=\begin{bmatrix}0 &0  \end{bmatrix}^{T} \label{eq:terminal_c}
\end{align}
where $\delta_l$ and $\delta_r$ are the distance at each global racing line waypoint to the left and the right track boundary, respectively. The terminal constraint \eqref{eq:terminal_c} is used to make sure the last two points of the ego trajectory are rejoining the global racing line ensuring a smooth transition between the planned and global trajectory. Based on previous analysis, the planning optimization problem is formulated as follows:
\begin{align} 
&\min_{d^{ego}_i}J(d^{ego})\\
&\;\text{s.t.}\;\;{d}_0^{ego}=d_{current}\notag\\
&\;\text{with}\;\;\eqref{eq:d_constr},\,\eqref{eq: radius_constr},\,\eqref{eq: bound_constr},\,
\eqref{eq:terminal_c} \notag\\
&\qquad\forall i=0,1,\cdots,N \notag 
\end{align}

\begin{table*}[htb]
    \centering
    \vspace{4.5mm}
    \resizebox{\textwidth}{!}
    {%
    \begin{tabular}{l|cc|cc|cc|cc}
    \toprule
    \textbf{Overtaking Planner} & \multicolumn{2}{|c}{\textbf{Racing Line}} & \multicolumn{2}{|c}{\textbf{Shortest Path}} & \multicolumn{2}{|c}{\textbf{Centerline}} & \multicolumn{2}{|c}{\textbf{Reactive}} \\
    \midrule
    &\bm{$\mathcal{R}_{ot/c}[\%]\uparrow$} & \bm{$\mathcal{S}_{max}[\%]\uparrow$}
    &\bm{$\mathcal{R}_{ot/c}[\%]\uparrow$} & \bm{$\mathcal{S}_{max}[\%]\uparrow$}
    &\bm{$\mathcal{R}_{ot/c}[\%]\uparrow$} & \bm{$\mathcal{S}_{max}[\%]\uparrow$}
    &\bm{$\mathcal{R}_{ot/c}[\%]\uparrow$} & \bm{$\mathcal{S}_{max}[\%]\uparrow$}\\
    \midrule
    
    Frenet    & \textbf{71.4} & 33.0    & 71.4 & 53.0   & 55.6 & 49.2 & \textbf{83.3} & 60.0 \\
    Graph Based   & 50.0 & 53.5   & 50.0 & 53.9   & \textbf{83.3} & 69.0   & 62.5 & 57.0 \\
    Spliner   & 55.5 & 81.2   & 55.5 & 64.5   & 62.5 & 71.9   & 55.5 & 62.2 \\
    Pred. Spliner \textbf{(ours)}  & \textbf{71.4} & \textbf{83.1}   & \textbf{100.0} & \textbf{69.7}  &  \textbf{83.3} & \textbf{80.2}   & \textbf{83.3} & \textbf{70.4} \\
    \midrule
    MPC                 & 50.0 & 80.7    & 55.5 & 56.1   & 41.7 & 70.8 & N.C. & 68.7 \\
    Pred. MPC (\textbf{ours})    & \textbf{71.4} & \textbf{83.3}    & \textbf{83.3} & \textbf{65.8}   & \textbf{83.3} & \textbf{82.2} & \textbf{83.3} & \textbf{82.4} \\
    \bottomrule
    \end{tabular}%
    }
    \caption{Overtaking performance of each evaluated planner across four different opponent behaviors on a 1:10 scale physical autonomous racing platform. The overtaking success rate ($\mathcal{R}_{ot/c}$) represents the percentage of successful overtakes relative to the number of attempts that resulted in a crash. The maximum speed scalar ($\mathcal{S}_{max}$) denotes the highest speed at which the overtaking planner can successfully overtake an opponent as a percentage of the ego vehicle's speed. \gls{mpc} based comparisons are evaluated separately, to isolate the overtaking algorithm comparison from the controller. N.C. denotes that the planner did not manage to perform 5 successful overtakes.}
    \label{tab:phys_res}
\end{table*}

The optimization problem is solved using an \gls{sqp} implemented in the \texttt{scipy.optimize} package. Providing the solver with a feasible initial guess for the overtaking side from the \emph{Spliner} output \cite{forzaeth} expedites the derivation of a solution. Once a solution has been found by the solver, it is used as the initial guess for the next iteration. Hence, the avoidance trajectory is continuously generated until the ego car successfully overtakes \rev{(i.e. $s_{ego} > s_{opp}$)}.

\begin{figure}[htb]
    \centering
    \includegraphics[width=\columnwidth]{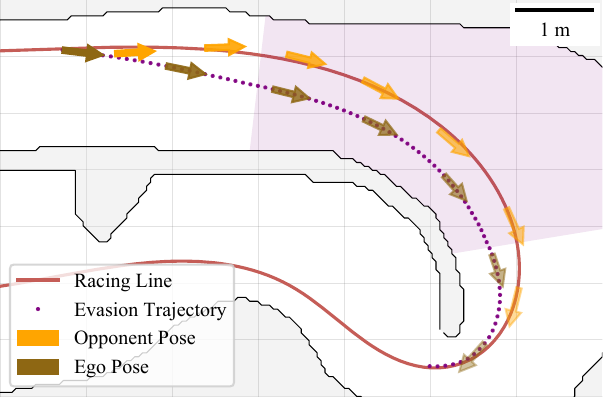}
    \caption{Exemplified trajectory at the starting instant of the overtake. The \gls{sqp}-based evasion trajectory, shown with purple dots, avoids the opponent within the shaded \gls{roc}. Poses of both the \emph{ego} agent and the \emph{opp} are shown with progressively shaded hues, to identify the progressing timesteps. At the end of the maneuver, the evasion trajectory reconnects the reference racing line, shown in red.}
    \label{fig:sqp}
\end{figure}

\rev{\Cref{fig:sqp} shows an example overtaking solution from the \gls{sqp}-based method. The trajectory is computed within the \gls{roc} defined by $c_{start}$ and $c_{end}$ and the track boundaries, allowing the planner to avoid suboptimal or premature overtakes seen in purely spatial planners (\cref{tab:overtaking_algorithms}). Using future opponent positions sampled from $\mathcal{GP}_d(s_{RoC})$ for $s_{RoC} \in [c_{start}, c_{end}]$, this approach incorporates both spatial and temporal information for safe, effective racing.}

\begin{figure*}[!htb]
    \includegraphics[width=\textwidth]{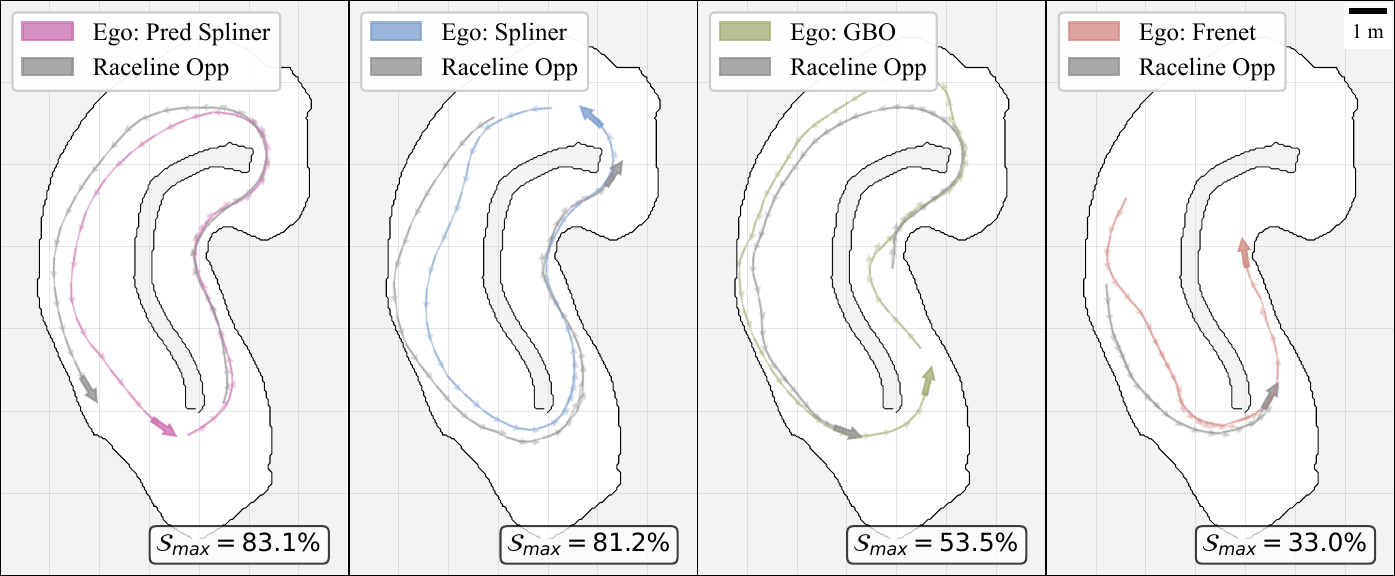}
    \caption{Overtake comparison corresponding to the results of Table \ref{tab:phys_res} from left to right: \emph{Predictive Spliner} (\textbf{ours}), \emph{Spliner} \cite{forzaeth}, \gls{gbo} \cite{gbo}, and \emph{Frenet} \cite{frenetplanner}. Each subplot illustrates the trajectory followed by the \emph{ego} agent (colored path) and the \emph{opponent} agent (gray path) during an overtaking maneuver. The maximum speed scaling $\mathcal{S}_{max}$ for each method indicates the highest speed at which the \emph{ego} agent can successfully overtake the opponent. In all settings, the opponent is tracking the optimal racing line at their respective maximum speed scaling $\mathcal{S}_{max}$. The \emph{Predictive Spliner} shows the highest $\mathcal{S}_{max}$, demonstrating its superior capability to overtake at higher speeds compared to the other methods.
} 
    \label{fig:qualitative_ots}
\end{figure*}

\begin{figure}[!htb]
    \includegraphics[width=\columnwidth]{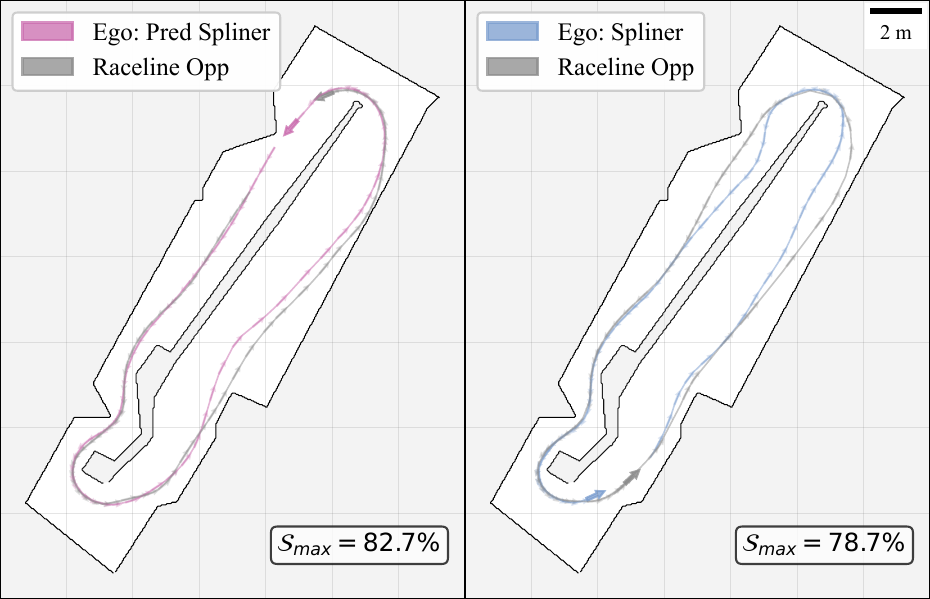}
    \caption{Left: An overtaking maneuver by the proposed \emph{Predictive Spliner}. Right: An overtaking attempt by the \emph{Spliner} method without temporal opponent prediction. In both settings, the opponent is tracking the optimal racing line, both at their respective maximum speed scaling $\mathcal{S}_{max}$ for the given map.} 
    \label{fig:qualitative_ots_etf}
\end{figure}

\section{Experimental Results}
This section presents the evaluation of the proposed \emph{Predictive Spliner} algorithm against various overtaking planners. The experimental setup is detailed in \Cref{sec:expsetup} and the results are evaluated in \Cref{subsec:phys_res}.

\subsection{Experimental Setup}
\label{sec:expsetup}
The proposed \emph{Predictive Spliner} algorithm was evaluated against established overtaking algorithms \cite{ftg, frenetplanner, gbo} on 1:10 scale physical systems \cite{forzaeth}. The \emph{ego} agent followed a global minimum curvature racing line \cite{gb_optimizer} optimized for maximum cornering speed, while the \emph{opp} agent's speed was set to the maximum value that still allowed overtaking, where the lap times $\mathcal{T}_{ego/opp}$ on an unobstructed track were used to compute said speed-scaler $\mathcal{S}:=\frac{\mathcal{T}_{opp}}{\mathcal{T}_{ego}} \in [0,1]$. In each experiment, the number of successful overtakes $\mathcal{N}_{ot}$ and crashes $\mathcal{N}_{c}$ were logged to compute the overtake success-rate $\mathcal{R}_{ot/c}:=\frac{\mathcal{N}_{ot}}{\mathcal{N}_{ot}+\mathcal{N}_{c}} \in [0,1]$. An experiment run lasted until 5 successful overtakes were achieved, hence $\mathcal{N}_{ot}=5$. The \emph{opp} agent's behavior was tested under four distinct settings. The different \emph{opp} behaviors are defined as follows:

\begin{enumerate}[I]
    \item \textbf{Racing Line:} The trajectory followed by the \emph{opp} agent is identical to that of the \emph{ego} agent, forcing the \emph{ego} to deviate from its optimal path. The \emph{opp} agent's behavior remains uninfluenced by the \emph{ego} agent. The racing line represents the global minimum curvature trajectory.
    
    \item \textbf{Shortest Path:} The racing line of the \emph{opp} agent follows the shortest path which overlaps with that of the \emph{ego} only at the apex of corners. The presence of the \emph{ego} agent does not alter the \emph{opp} agent's behavior.
    
    \item \textbf{Centerline Trajectory:} The centerline of the race track is the designated path for the \emph{opp} agent, diverging considerably from the \emph{ego} agent's racing line. This behavior is also maintained regardless of the \emph{ego} agent's maneuvers.
    
    \item \textbf{Reactive Opponent:} A reactive control strategy is adopted by the \emph{opp} agent, guided by the longest path observed through its \gls{lidar} sensor \cite{ftg}. This strategy does not follow a predetermined racing line. While generally less performant, this results in unpredictable movements influenced by the overtaking actions of the \emph{ego} agent.
\end{enumerate}

\subsection{Physical Results} \label{subsec:phys_res}
As shown in \Cref{tab:phys_res}, the proposed \emph{Predictive Spliner} overtaking algorithm achieves the highest $\mathcal{S}_{max}$ scalar across all four different opponent behaviors. This means it can overtake the fastest opponents, of up to 83.1\% of its ego speed. Additionally, \emph{Predictive Spliner} demonstrates the highest overtaking success rate $\mathcal{R}_{ot/c}$, indicating that it is the safest method for overtaking. Consequently, it can overtake the fastest opponents while maintaining the highest success rate for its maneuvers. Compared to the second most performant comparison method, \emph{Spliner} \cite{forzaeth}, \emph{Predictive Spliner} can on average --- across the four different opponent behaviors --- overtake opponents that are 8.4\% faster ($\bar{\mathcal{S}}_{max,PSpliner}=75.85$\%, $\bar{\mathcal{S}}_{max,spliner}=69.95$\%) while maintaining a 47.6\% higher success rate ($\bar{\mathcal{R}}_{ot/c,PSpliner}=84.50$\%, $\bar{\mathcal{R}}_{ot/c,spliner}=57.25$\%). \rev{Notably, when subjected to a reactive opponent whose trajectory is non-deterministic (i.e. the \(\mathcal{GP}\) assumption that the leading vehicle maintains a fixed
trajectory could be violated), in practice, the proposed method still manages to significantly outperform the \emph{Spliner} method in terms of safety and speed by more than 8\% points.} \rev{Lastly, these results align with the method's real-world performance, as \emph{Predictive Spliner} has been successfully employed in the \emph{ICRA and IROS 2024} F1TENTH Grand Prix, demonstrating its effectiveness and robustness under competitive conditions.
}


\rev{\Cref{tab:phys_res} presents \gls{mpc}-based overtaking separately to isolate the impact of predictiveness on the planner from the controller's performance, as other strategies use the \gls{map} controller \cite{forzaeth} for faster tracking ($\mathcal{T}_{ego,MAP}=6.9s$, $\mathcal{T}_{ego,MPC}=7.1s$, \cref{fig:qualitative_ots}). This demonstrates that predictiveness-enhanced overtaking can improve any strategy’s performance. While standard \gls{mpc} applies boundary constraints around immediate opponent detections, \emph{Predictive} \gls{mpc} uses future opponent constraints within the \gls{roc}, allowing for a straightforward integration of the \gls{gp} predictions. Predictiveness increased the average $\bar{\mathcal{R}}_{ot/c}$ from 49.1\% to 80.3\% (+63.5\%), though our primary focus remains on \emph{Predictive Spliner}.}

\Cref{fig:qualitative_ots} qualitatively demonstrates the results of \Cref{tab:phys_res} and shows how incorporating the opponent's future behavior allows for more effective overtaking performance. It shows how the proposed \emph{Predictive Spliner} can leverage the opponent's information and yields the most direct overtaking trajectory of the four evaluated overtaking strategies.


\rev{\Cref{fig:qualitative_ots_etf} provides an additional qualitative comparison between \emph{Spliner} and \emph{Predictive Spliner} on a second track. As before, the \emph{Spliner} method \cite{forzaeth} exhibits more premature overtaking attempts, especially against slower opponents, limiting its success at higher speeds. In contrast, \emph{Predictive Spliner} effectively predicts opponent behavior, reducing premature attempts and thus enabling successful overtakes at higher speeds.}

\subsection{Computational Results}
As shown in \Cref{tab:compute}, the proposed \textit{Predictive Spliner} overtaking planner demonstrates competitive computational efficiency with a computational time of \SI{8.4}{\milli\second}, \SI{22.79}{\percent} \gls{cpu} load, and \SI{85.59}{\mega\byte} memory usage. Although these computational demands are not the lowest among the compared algorithms, they remain relatively low. Considering the significant performance improvement highlighted in \Cref{tab:phys_res}, the computational overhead with its \SI{2.9}{\milli\second} latency increase, of the \textit{Predictive Spliner} planner can be deemed acceptable. Note that the \gls{cpu} load is measured using the \texttt{psutil} tool. In this context, 100\% utilization represents the full load of a single core. Therefore, on a multi-core \gls{cpu} such as the utilized Intel i7-1165G7, the total utilization can exceed 100\%. Further, \gls{mpc} computation times can not be compared with the planners, as \gls{mpc} performs planning and controls simultaneously.

\begin{table}[htb] 
    \centering 
    \resizebox{\columnwidth}{!}{%
    \begin{tabular}{l|cc|cc|cc}
    \toprule
    \textbf{Overtaking Planner} & \multicolumn{2}{c|}{\textbf{\gls{cpu} [\%]$\downarrow$}} & \multicolumn{2}{c|}{\textbf{Mem [MB]}$\downarrow$} & \multicolumn{2}{c}{\textbf{Comp. Time [ms]}$\downarrow$} \\
    & $\mu_{cpu}$ & $\sigma_{cpu}$ & $\mu_{mem}$ & $\sigma_{mem}$ & $\mu_{t}$ & $\sigma_{t}$ \\
    \midrule
    Frenet & 36.74 & 12.78 & \textbf{41.04} & 2.11 & 10.78 & 38.38 \\
    Graph Based & 98.97 & \textbf{3.27} & 170.41 & \textbf{0.01} & 40.13 & 9.20 \\
    Spliner & \textbf{16.74} & 4.25 & 59.92 & 0.14 & \textbf{5.50} & \textbf{2.74 }\\
    Pred. Spliner \textbf{(ours)} & 22.79 & 31.19 & 85.59 & 0.81 & 8.40 & 26.64 \\
    \midrule
    MPC & 106.93 & 2.00 & 97.84 & 0.01 & 27.52 & 5.44\\
    Pred. MPC \textbf{(ours)} & 106.76 & 2.47 & 99.79 & 0.21 & 64.61 & 9.79 \\
    \bottomrule
    \end{tabular}%
    }
    \caption{Computational results of each overtaking strategy. \gls{cpu} load and memory usage using the \texttt{psutil} tool. The \gls{cpu} load, memory usage, and computational times are reported with their mean $\mu$ and standard deviation $\sigma$ and are all computed on an Intel i7-1165G7.}
    \label{tab:compute}
\end{table}

\section{Conclusion}
We present \emph{Predictive Spliner}, a data-driven overtaking algorithm for autonomous racing that learns and leverages opponent behavior via \gls{gp} regression. This method \rev{reduces} premature overtaking attempts and successfully overtakes opponents at up to 83.1\% of its speed, with an average success rate of 84.5\%, outperforming previous methods by 47.6\%\rev{, while maintaining low computational demands for real-time applications. Validated on a 1:10 scale platform and deployed in real competitive F1TENTH races (\emph{ICRA and IROS 2024}), \emph{Predictive Spliner} demonstrates practical applicability.} \rev{Future work could focus on combining game-theoretical approaches with similar methods from \emph{Predictive Spliner}, or on handling multiple opponents. The same principles can be applied, given the feasibility of multi-object tracking, and in such cases, the \gls{roc} computation can then be adapted to iterate over all available opponent trajectories.}







\bibliographystyle{IEEEtran}
\bibliography{main}

\end{document}